\documentclass[letterpaper, 10 pt, journal, twoside]{IEEEtran}

\usepackage[T1]{fontenc}
\usepackage{times}
\usepackage{amsmath}
\usepackage{amssymb}
\usepackage{graphicx}
\usepackage[ruled,linesnumbered]{algorithm2e}
\usepackage{booktabs}
\usepackage{float}
\usepackage{bigstrut}
\usepackage{multirow}
\usepackage{balance}
\usepackage[hidelinks=true,bookmarks=FWLse]{hyperref}
\usepackage{lineno}
\usepackage{amsopn}
\usepackage{comment}
\usepackage{color}
\usepackage{cite}
\usepackage{algorithmic}
\usepackage{tabularx}
\usepackage{xcolor}
\usepackage{mathrsfs}
\usepackage{makecell}

\usepackage{url,subfigure,tabulary}

\newcommand{\ie}{\textit{i.e.}}

\newcommand{\etal}{\textit{et al.}}

\usepackage{array}
\makeatletter
\newcommand{\thickhline}{%
    \noalign {\ifnum 0=`}\fi \hrule height 1pt
    \futurelet \reserved@a \@xhline
}
\newcolumntype{"}{@{\hskip\tabcolsep\vrule width 1pt\hskip\tabcolsep}}
\makeatother

\makeatletter

\newcommand{\Rmnum}[1]{\expandafter\@slowromancap\romannumeral #1@}
\makeatother

\newcolumntype{L}[1]{>{\raggedright\arraybackslash}p{#1}}
\newcolumntype{C}[1]{>{\centering\arraybackslash}p{#1}}
\newcolumntype{R}[1]{>{\raggedleft\arraybackslash}p{#1}}

\title{
    Dynamic Fusion Module Evolves\\ Drivable Area and Road Anomaly Detection:\\ A Benchmark and Algorithms
}

\author{Hengli~Wang,~\IEEEmembership{Graduate Student Member,~IEEE},~Rui~Fan,~\IEEEmembership{Member,~IEEE},\\~Yuxiang~Sun,~\IEEEmembership{Member,~IEEE},~and~Ming~Liu,~\IEEEmembership{Senior Member, IEEE}
\thanks{\textit{(Corresponding author: Ming Liu.)}}
\thanks{H. Wang and M. Liu are with the Department of Electronic and Computer Engineering, the Hong Kong University of Science and Technology, Clear Water Bay, Kowloon, Hong Kong SAR, China (email: hwangdf@connect.ust.hk; eelium@ust.hk).}
\thanks{R. Fan is with the Department of Computer Science and Engineering, and the Department of Ophthalmology, the University of California San Diego, La  Jolla, CA 92093, United States (email: rui.fan@ieee.org).}
\thanks{Y. Sun is with the Department of Mechanical Engineering, The Hong Kong Polytechnic University, Hung Hom, Kowloon, Hong Kong (e-mail:
yx.sun@polyu.edu.hk, sun.yuxiang@outlook.com).}
\thanks{H. Wang and R. Fan contributed equally to this work.}
}

\begin{document}

\maketitle

\begin{abstract}
    Joint detection of drivable areas and road anomalies is very important for mobile robots. Recently, many semantic segmentation approaches based on convolutional neural networks (CNNs) have been proposed for pixel-wise drivable area and road anomaly detection. In addition, some benchmark datasets, such as KITTI and Cityscapes, have been widely used. However, the existing benchmarks are mostly designed for self-driving cars. There lacks a benchmark for ground mobile robots, such as robotic wheelchairs. Therefore, in this paper, we first build a drivable area and road anomaly detection benchmark for ground mobile robots, evaluating the existing state-of-the-art single-modal and data-fusion semantic segmentation CNNs using six modalities of visual features. Furthermore, we propose a novel module, referred to as the dynamic fusion module (DFM), which can be easily deployed in existing data-fusion networks to fuse different types of visual features effectively and efficiently. The experimental results show that the transformed disparity image is the most informative visual feature and the proposed DFM-RTFNet outperforms the state-of-the-arts. Additionally, our DFM-RTFNet achieves competitive performance on the KITTI road benchmark. Our benchmark is publicly available at \url{https://sites.google.com/view/gmrb}.
\end{abstract}


\section{Introduction}
\label{sec.introduction}
\IEEEPARstart{M}{obile} robots are becoming increasingly popular in our daily life thanks to their benefits. Self-driving cars, for example, can greatly reduce traffic accidents \cite{chai2020multiobjective,liu2021the,sun2021pointmoseg}. Ground mobile robots, such as robotic wheelchairs and sweeping robots, can significantly improve people's comfort and life quality~\cite{li2018adaptive,wang2019self,wang2020applying}. Visual environmental perception and autonomous navigation are two fundamental components in mobile robots. The former analyzes the input sensory data and outputs environmental perception results, with which the latter enables the robot to autonomously move from its current position to its destination \cite{sun2019accurate,sun2020see,wang2021s2p2}. Among all visual environmental perception tasks for mobile robots, the joint detection of drivable areas and road anomalies at the pixel-level is a crucial one. Accurate and efficient drivable area and road anomaly detection can help avoid accidents for such vehicles. Note that in this paper, drivable area refers to a region in which mobile robots can navigate safely, while a road anomaly refers to a region with a height difference from the surface of the drivable area.

With the rapid development of deep learning technologies, many semantic segmentation approaches based on convolutional neural networks (CNNs) have been proposed, and these methods can be used for drivable area and road anomaly detection. For example, Chen \etal \cite{chen2018encoder} proposed DeepLabv3+, which combines the spatial pyramid pooling (SPP) module and the encoder-decoder architecture to generate semantic predictions. Recently, data-fusion networks have been proposed to improve the detection performance by extracting and fusing two different types of visual features. Specifically, Wang \etal \cite{wang2018depth} proposed a depth-aware operation to fuse RGB and depth images. In addition, Zhang \etal \cite{zhang2017multi} fused RGB images with elevation maps. All of these fusion strategies demonstrated the superior performance using multi-modal data.

\begin{figure*}[t]
    \centering
    \includegraphics[width=\textwidth]{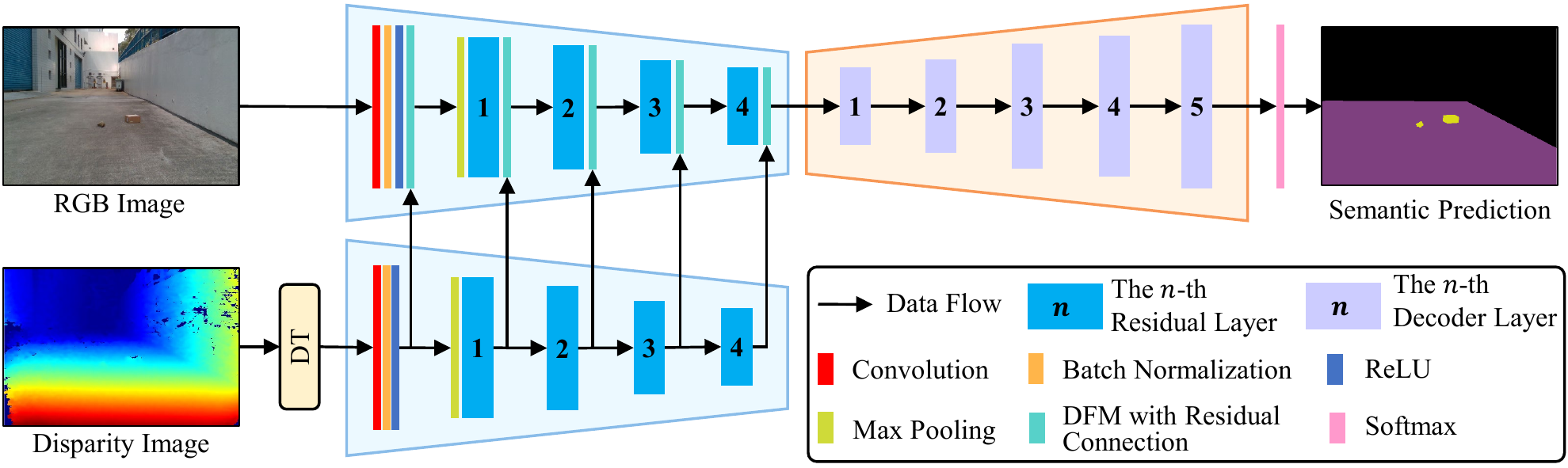}
    \caption{An illustration of our drivable area and road anomaly detection framework. It consists of disparity transformation (DT), two encoders (the blue trapezoids) and one decoder (the orange trapezoid). DT first transforms the input disparity image. Then, our DFM dynamically fuses two different modalities of features in a multi-scale fashion. Finally, the fused feature is processed by five decoder layers and a softmax layer to output the detection result.}
    \label{fig.network}
\end{figure*}

\subsection{Motivation and Novel Contributions}
\label{sec.motivation_and_novel_contributions}
Different types of mobile robots usually work in different environments, so they usually have different focuses on environmental perception. Self-driving cars, for example, focus mostly on cars and pedestrians because they run on roadways. Differently, ground mobile robots need to pay more attention to small obstacles, such as stones and tin cans, since they usually work in indoor environments or on outdoor sidewalks. Unfortunately, current public semantic segmentation benchmarks, such as KITTI \cite{geiger2012we} and Cityscapes \cite{cordts2016cityscapes}, focus mainly on driving scenarios. Other scenarios, such as robotic wheelchairs running on sidewalks, are not included. Therefore, in this paper, we first build a drivable area and road anomaly detection benchmark for ground mobile robots using our previously published ground mobile robot perception (GMRP) dataset\footnote{\url{https://github.com/hlwang1124/GMRPD}} \cite{wang2019self}, on which the performances of state-of-the-art (SOTA) single-modal and data-fusion networks using different types of training data are compared in detail.

Recently, incorporating different modalities of visual features into semantic segmentation has become a promising research direction that deserves more attention \cite{fan2020sne}. The visual features such as depth and elevation can greatly improve the detection performance. Our recent work \cite{fan2019pothole} introduced a new type of visual feature, referred to as the transformed disparity image, in which the values of drivable areas become similar but the value differences between drivable areas and road anomalies/damages become significant. This can help discriminate drivable areas and road anomalies.

Moreover, the existing data-fusion networks typically fuse two different modalities of data by performing simple element-wise addition or feature concatenation. However, we can improve the detection performance by using more effective fusion operations. Inspired by the dynamic filtering network~\cite{jia2016dynamic}, we propose a novel data-fusion module, named the dynamic fusion module (DFM), as shown in Fig.~\ref{fig.network}, which can be easily deployed in the existing data-fusion semantic segmentation networks to dynamically generate the fused feature representations using content-dependent and spatially variant kernels.

In the experiments, we first compare our proposed network, referred to as DFM-RTFNet, with ten SOTA single-modal networks and four SOTA data-fusion networks, using six different types of training data: (a) RGB images, (b) disparity images, (c) normal images, (d) HHA images, (e) elevation maps and (f) transformed disparity images. The experimental results demonstrate that the transformed disparity image is the most informative visual feature, and it can greatly improve the drivable area and road anomaly detection performance. Moreover, our DFM-RTFNet achieves the best overall performance. We further evaluate our DFM-RTFNet on the KITTI benchmark to validate its effectiveness for self-driving cars, and the experimental results illustrate that our DFM-RTFNet achieves competitive performance on the KITTI road benchmark \cite{fritsch2013new}.

\subsection{Paper Outline}
\label{sec.paper_outline}
The remainder of this paper is organized as follows: Section~\ref{sec.related_work} reviews related work. Section~\ref{sec.disparity_transformation} introduces our previously published disparity transformation algorithm used in this paper. In Section~\ref{sec.dynamic_fusion_module}, we introduce our DFM-RTFNet. Section~\ref{sec.experiment_results_and_discussion} presents the experimental results and compares our framework with other SOTA approaches. Finally, we conclude this paper in the last section.

\section{Related Work}
\label{sec.related_work}
In this section, we first overview some selected SOTA single-modal and data-fusion networks for semantic segmentation. We also compare these networks with our DFM-RTFNet in Section~\ref{sec.drivable_area_and_road_anomaly_detection_benchmark} and \ref{sec.evaluations_on_the_kitti_benchmark}. Secondly, we introduce several visual features acquired from 3D geometry information. Finally, we briefly review the dynamic filtering techniques.

\subsection{SOTA Networks for Semantic Segmentation}
\label{sec.soa_networks_for_semantic_segmentation}
FCN \cite{long2015fully} was the first end-to-end single-modal approach for semantic segmentation. It employs an image classification network for feature extraction with the fully connected layers removed. There are three main FCN variants: FCN-32s, FCN-16s and FCN-8s, and we use FCN-8s in our experiments.

SegNet \cite{badrinarayanan2017segnet} was the first to present the encoder-decoder architecture, which is widely used in current networks. It consists of an encoder network, a corresponding decoder network and a final pixel-wise classification layer. U-Net \cite{ronneberger2015u} is designed based on the concept of the encoder-decoder architecture, and adds skip connections between the encoder and decoder to help restore the location of small objects.

PSPNet \cite{zhao2017pyramid} was the first to employ a pyramid pooling module to extract useful context information for better performance. DeepLabv3+ \cite{chen2018encoder} follows this idea and employs depth-wise separable convolution to both atrous SPP (ASPP) and the decoder module. DenseASPP \cite{yang2018denseaspp} further connects a set of atrous convolutional layers in a dense way.

To further improve the performance, UPerNet \cite{xiao2018unified} tries to identify many visual concepts such as objects and textures in parallel; DUpsampling \cite{tian2019decoders} exploits a data-dependent decoder that considers the correlation among the prediction of each pixel; and GSCNN \cite{takikawa2019gated} utilizes a novel architecture consisting of a shape branch and a regular branch, which can help each other focus on the relevant boundary information. Moreover, ESPNet \cite{mehta2018espnet} decomposes the standard convolution layer to save memory and computation cost.

As previously mentioned, data-fusion networks have been proposed to improve detection performance. Such architectures generally use two different types of visual features to learn informative representations. Specifically, FuseNet \cite{hazirbas2016fusenet} adopts the encoder-decoder architecture, and employs element-wise addition to fuse the feature maps of the RGB branch and depth branch. Different from FuseNet, the depth-aware CNN \cite{wang2018depth} introduces two novel operations, depth-aware convolution and depth-aware average pooling. These operations can leverage depth similarity between pixels to incorporate geometric information into the CNN. RTFNet~\cite{sun2019rtfnet} was developed to enhance the performance of semantic segmentation using RGB images and thermal images. It also adopts the encoder-decoder design concept and element-wise addition fusion strategy. Moreover, MFNet \cite{ha2017mfnet} was developed to fuse RGB images and thermal images for real-time operation.

However, these data-fusion networks often fuse two different types of information by performing simple element-wise addition or feature concatenation. We think that it is difficult to fully exploit two different modalities of data in such a simple fusion way. Different from previous work, our proposed DFM can dynamically generate the fused feature representations using content-dependent and spatially variant kernels, which can significantly improve the detection performance.

\subsection{Visual Features Acquired From 3D Geometry Information}
\label{sec.visual_features}
Many researchers have proposed visual features computed using 3D geometry information to improve the drivable area and road anomaly detection performance. Specifically, Gupta \etal \cite{gupta2015indoor} combined RGB images with normal information for multiple tasks including contour classification and semantic segmentation. Zhang \etal \cite{zhang2017multi} fused RGB images with elevation maps to improve the semantic segmentation performance. In addition, HHA images were developed to act as complementary information for RGB images \cite{hazirbas2016fusenet}. An HHA image has three channels: (a) disparity, (b) height of the pixels and (c) the angle between the normals and the gravity vector based on the estimated ground floor. In \cite{fan2019pothole, fan2021tcyb}, we proposed a novel visual feature, referred to as the transformed disparity image, in which the drivable areas and road anomalies are highly distinguishable. The corresponding performance comparison is presented in Section~\ref{sec.drivable_area_and_road_anomaly_detection_benchmark} and \ref{sec.further_discussion}. 

\subsection{Dynamic Filtering Techniques}
\label{sec.dynamic_filtering_techniques}
The dynamic filtering network (DFN) \cite{jia2016dynamic} initially implemented the concept of dynamic filtering for the video prediction task, where the filters are dynamically generated based on one frame to process another frame. Recently, several extensions of the DFN have been proposed. For example, Simonovsky \etal \cite{simonovsky2017dynamic} extended the DFN for graph classification. Wu~\etal \cite{wu2018dynamic} developed an extension of the DFN by dynamically generating weights to enlarge receptive fields for flow estimation. Our proposed DFM can also be regarded as an extension of the DFN for data-fusion semantic segmentation. We adopt the same philosophy as DFN, and make specific adjustments to save GPU memory for multi-scale feature fusion, which has not been studied by previous papers.

\begin{figure}[t]
    \centering
    \includegraphics[width=0.485\textwidth]{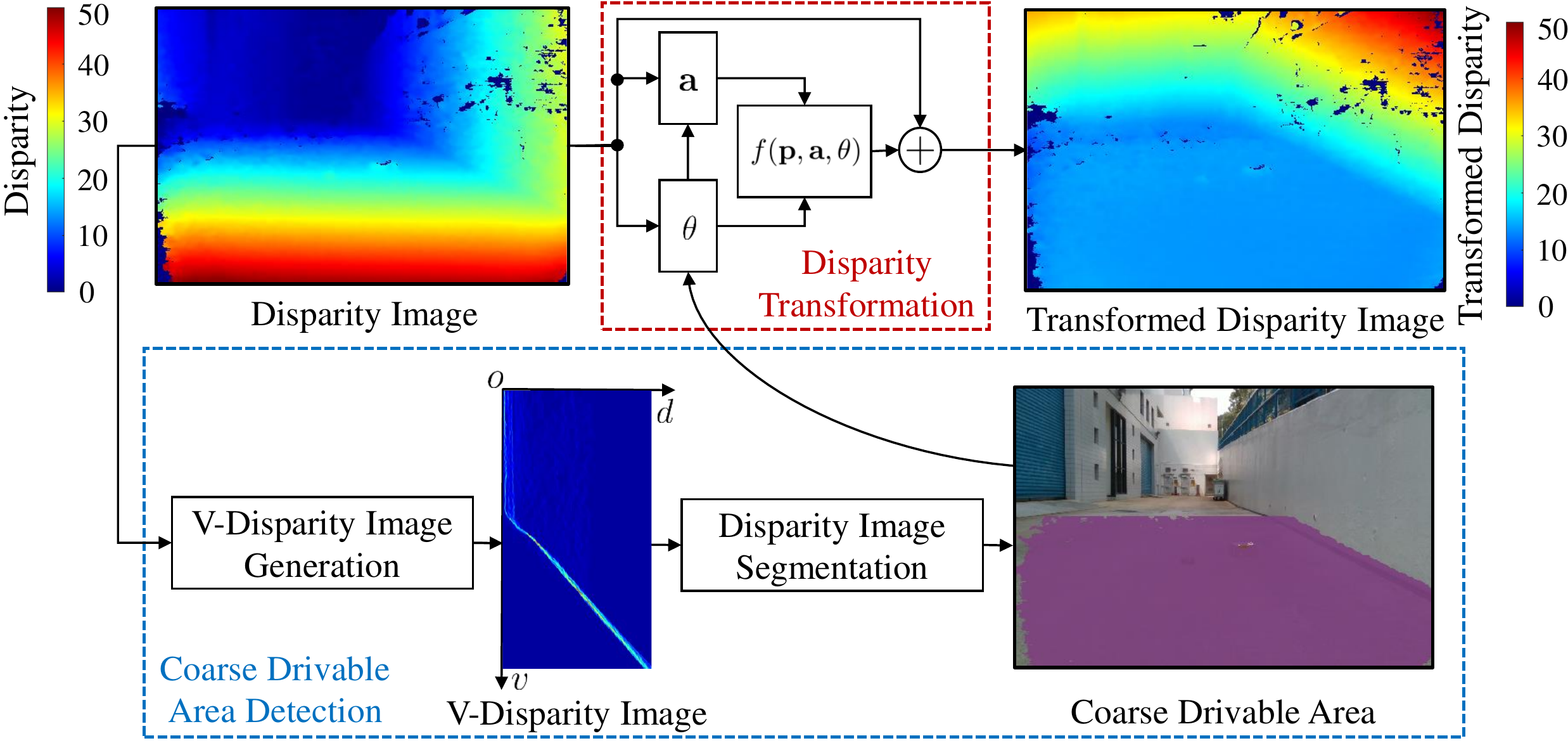}
    \caption{An illustration of transformed disparity image generation and coarse drivable area detection.}
    \label{fig.DT}
\end{figure}

\begin{figure*}[t]
    \centering
    \includegraphics[width=\textwidth]{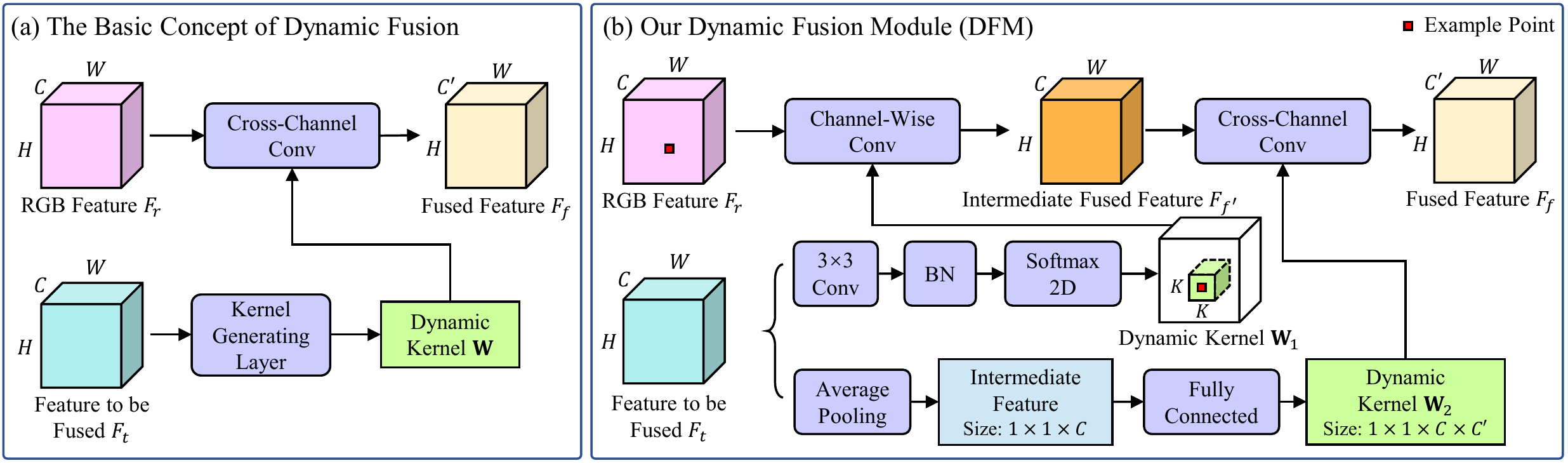}
    \caption{An illustration of the basic concept of dynamic fusion and our proposed DFM. (a) The basic concept of dynamic fusion is to produce a dynamic kernel $\mathbf{W}$ based on the feature to be fused $F_t$, and then apply $\mathbf{W}$ on the RGB feature $F_r$ to output the fused feature $F_f$. (b) Different from (a), our DFM adopts a two-stage convolution factorization process to save memory and computational resources.}
    \label{fig.DFM}
\end{figure*}

\section{Disparity Transformation}
\label{sec.disparity_transformation}
DT \cite{fan2019road} aims at transforming a disparity image into a quasi bird's eye view, so that the pixels of drivable areas possess similar values, while they differ greatly from those of the road damages/anomalies \cite{fan2020we}. The expression of DT is as follows:
\begin{equation}
    \mathbf{D}_\text{t} = \mathbf{D}_\text{o} - f(\mathbf{p},\mathbf{a},\theta) + \delta,
\label{eq.dt}
\end{equation}
where $\mathbf{D}_\text{o}$ is the original disparity image; $\mathbf{D}_\text{t}$ is the transformed disparity image; $f$ is a non-linear function representing the disparities in the drivable area; $\mathbf{p} =(u;v)$ is an image pixel; $\mathbf{a} =(a_0;a_1)$ stores the road profile model coefficients; $\theta$ is the stereo rig roll angle; and $\delta$ is a constant set to ensure that the transformed disparity values are non-negative. $\theta$ can be obtained by minimizing \cite{fan2021learning}
\begin{equation}
    E(\theta)
    =\mathbf{d}^\top\mathbf{d}-\mathbf{d}^\top\mathbf{T}(\theta)
    \bigg(\mathbf{T}(\theta)^\top\mathbf{T}(\theta)\bigg)^{-1}
    \mathbf{T}(\theta)^\top\mathbf{d},
    \label{eq.E}
\end{equation}
where $\mathbf{T}=[\mathbf{1}_k,\mathbf{v}\cos\theta-\mathbf{u}\sin\theta]$; $\mathbf{d}=(d_1;\cdots;d_k)$ is a $k$-entry vector of disparities; $\mathbf{u}=(u_1;\cdots;u_k)$ is a $k$-entry vector of horizontal coordinates; and $\mathbf{v}=(v_1;\cdots;v_k)$ denotes a $k$-entry vector of vertical coordinates. $\mathbf{a}$ can be estimated using \cite{fan2019road}
\begin{equation}
    \mathbf{a}(\theta)=\bigg(\mathbf{T}(\theta)^\top\mathbf{T}(\theta)\bigg)^{-1}
    \mathbf{T}(\theta)^\top\mathbf{d}.
    \label{eq.alpha}
\end{equation}

In this paper, we first utilize the road segmentation approach proposed in \cite{ozgunalp2016multiple} to detect a coarse drivable area through v-disparity image analysis and disparity image segmentation. The disparities in the detected coarse drivable area are then used to estimate $\mathbf{a}$ and $\theta$ for disparity image transformation, as shown in Fig.~\ref{fig.DT}.

\section{Dynamic Fusion Module}
\label{sec.dynamic_fusion_module}
In this section, we first introduce our proposed DFM, which can generate fused features effectively and efficiently, as illustrated in Fig.~\ref{fig.DFM}. Then, we explain how to employ our DFM in data-fusion networks for semantic segmentation, and introduce our DFM-RTFNet, as shown in Fig.~\ref{fig.network}.

Given an RGB feature $F_r$ of size $H \times W \times C$ and a feature to be fused $F_t$ of the same size, the basic concept of dynamic fusion is to produce a dynamic kernel $\mathbf{W}$ based on $F_t$ and then apply $\mathbf{W}$ on $F_r$ to output the fused feature $F_f$ of $H \times W \times C'$, as shown in Fig.~\ref{fig.DFM} (a). This process can be regarded as an extension of the dynamic filtering network \cite{jia2016dynamic} for data-fusion semantic segmentation, and has the following formulation:
\begin{equation}
    F_f=\mathbf{W}(F_t;\Omega) \otimes F_r,
\end{equation}
where $\Omega$ denotes the parameters of the kernel generating layer shown in Fig.~\ref{fig.DFM} (a); and $\otimes$ denotes cross-channel convolution. The dynamic kernel $\mathbf{W}$ has two properties: content dependence and spatial variance. The former means that $\mathbf{W}$ is based on the feature to be fused $F_t$, while the latter means that different kernels are employed to different spatial positions of the RGB feature $F_r$. These novel properties enable the network to apply appropriate kernels to different image regions, which can generate more effective fused features $F_f$ and thus improve the overall detection performance.

However, generating and applying the dynamic kernel in such a simple way could consume much memory and computational resources, which makes this idea hard to deploy in practice. To address this problem, we adopt the philosophy of MobileNets \cite{howard2017mobilenets} and design a two-stage convolution factorization process for our DFM, which can save significant computational resources, as shown in Fig.~\ref{fig.DFM} (b).

In the first stage, we generate a dynamic kernel $\mathbf{W}_1$, which is then applied on the RGB feature $F_r$ using the channel-wise convolution operation to output an intermediate fused feature $F_{f'}$. Specifically, one channel of $F_r$ is convolved with the corresponding channel of $\mathbf{W}_1$. The first stage can be formulated as follows:
\begin{equation}
    F_{f'}=\mathbf{W_1}(F_t;\Omega_1) \odot F_r,
\end{equation}
where $\Omega_1$ denotes the parameters of the kernel-generating operations shown in Fig.~\ref{fig.DFM} (b); and $\odot$ denotes channel-wise convolution. Note that these channel-wise convolutions are still spatially variant. Specifically, for an example point of $F_r$, we take the convolution kernel of $K \times K$ size at the corresponding position in $\mathbf{W}_1$ to conduct the channel-wise convolution, as illustrated in Fig.~\ref{fig.DFM} (b).

In the second stage, we employ an average pooling layer and a fully connected layer to generate a dynamic kernel $\mathbf{W}_2$, which is then applied on $F_{f'}$ to generate the fused feature $F_f$. The second stage has the following formulation:
\begin{equation}
    F_{f}=\mathbf{W_2}(F_{t};\Omega_2) \otimes F_{f'},
\end{equation}
where $\Omega_2$ denotes the parameters of the fully connected layer. This two-stage process can significantly improve the efficiency and make this idea feasible in practice.

Then, we implement our DFM-RTFNet by integrating DFM into RTFNet50 \cite{sun2019rtfnet}, as shown in Fig.~\ref{fig.network}. Specifically, we use our DFMs with residual connections to replace the original element-wise addition layers, and our DFMs dynamically fuse two different modalities of features in a multi-scale fashion. The fused feature is then processed by five decoder layers and a softmax layer sequentially to output the final detection result. For details of the network architecture, we refer readers to \cite{sun2019rtfnet}. Additionally, the number of fused feature channels is identical to the number of input feature channels, \ie, $C=C'$, and we set the size of the convolution kernel $K=3$. The determination of different fusion strategies of our DFM-RTFNet will be discussed in Section~\ref{sec.ablation_study}.

\begin{figure*}[!h]
    \centering
    \includegraphics[width=\textwidth]{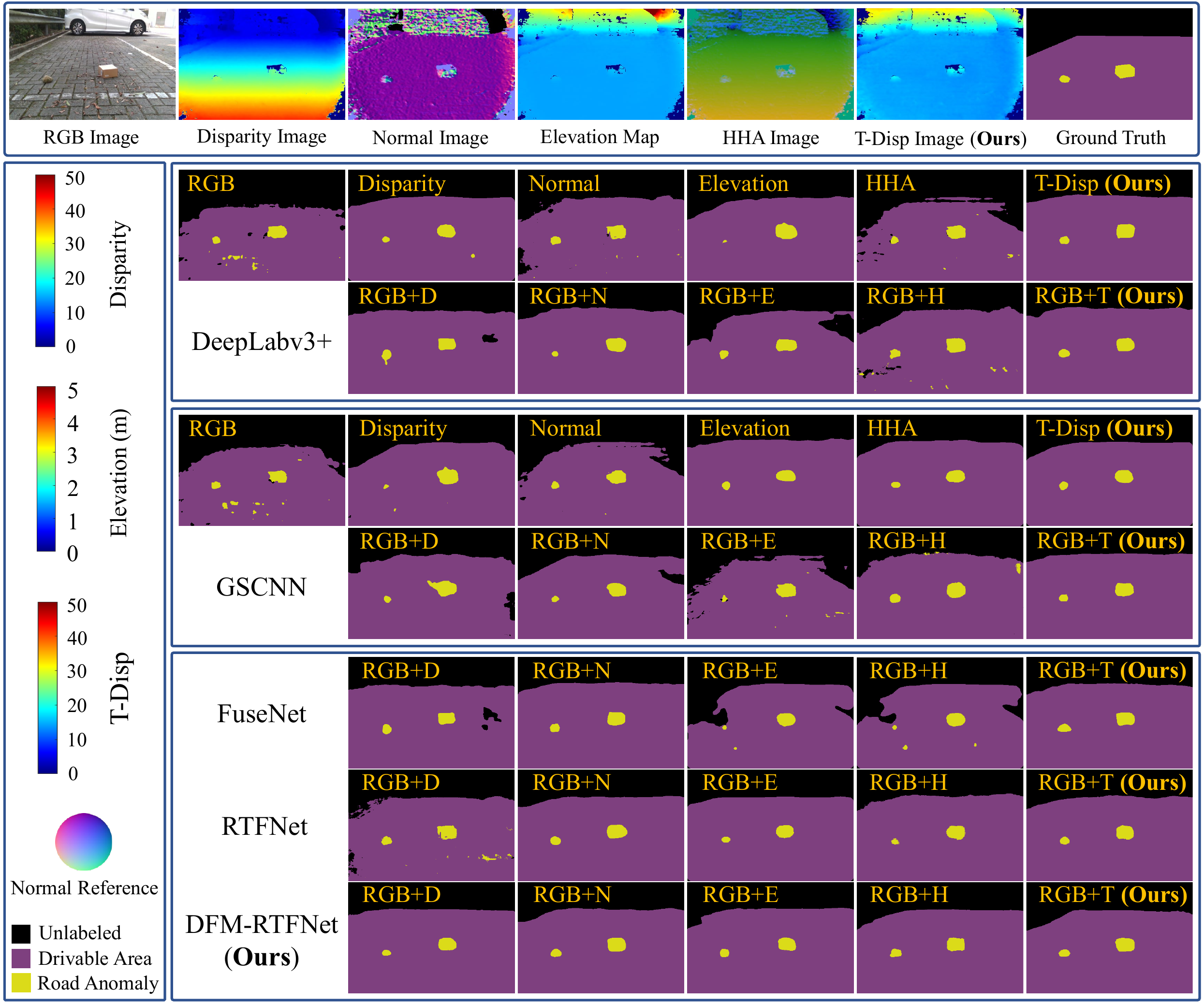}
    \caption{An example of the experimental results on our GMRP dataset. DeepLabv3+ \cite{chen2018encoder} and GSCNN \cite{takikawa2019gated} are single-modal networks, while FuseNet \cite{hazirbas2016fusenet}, RTFNet \cite{sun2019rtfnet} and our DFM-RTFNet are data-fusion networks.}
    \label{fig.benchmark}
\end{figure*}

\begin{figure*}[t]
    \centering
    \includegraphics[width=\textwidth]{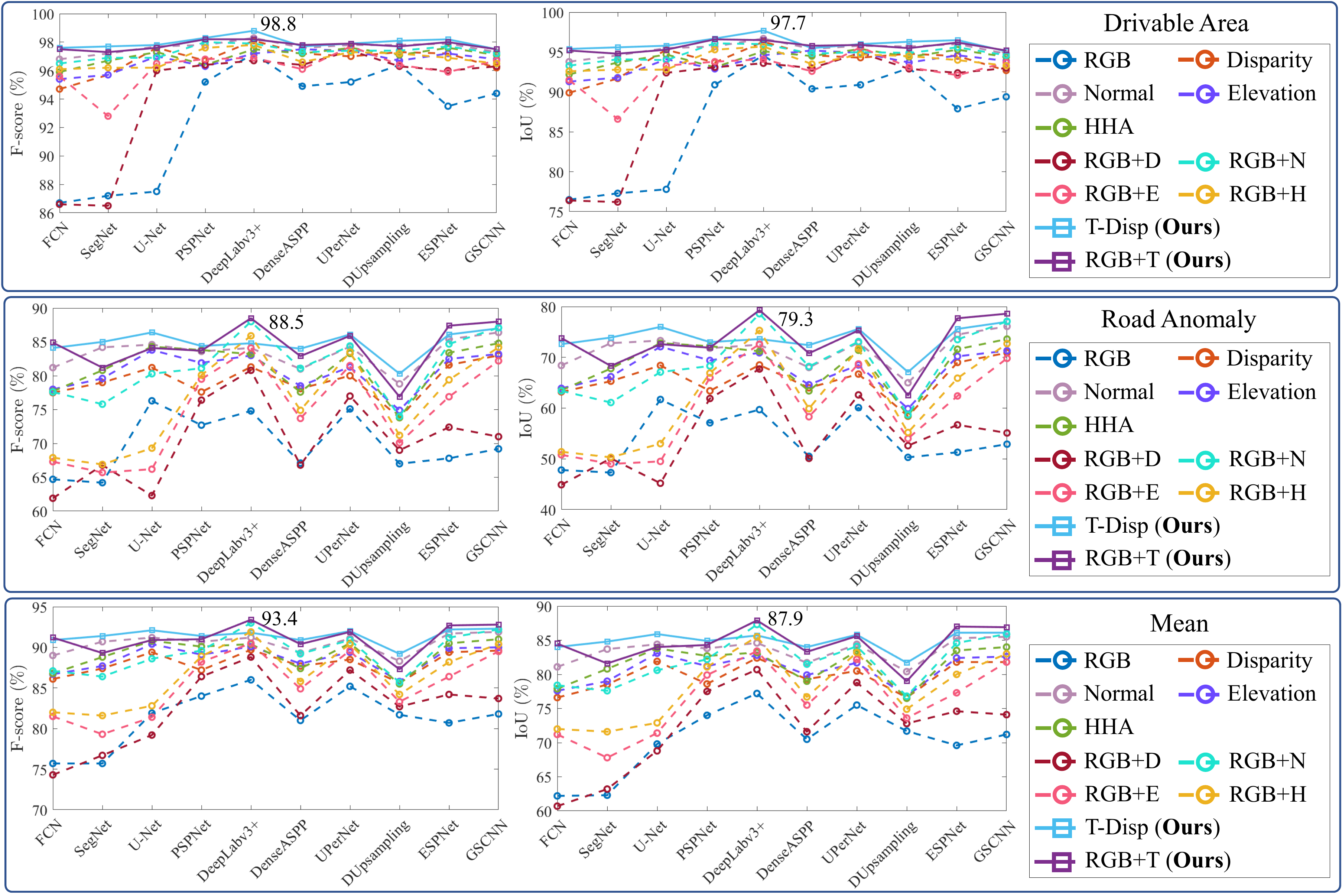}
    \caption{The performance comparison among ten SOTA single-modal networks (FCN \cite{long2015fully}, SegNet \cite{badrinarayanan2017segnet}, U-Net \cite{ronneberger2015u}, PSPNet \cite{zhao2017pyramid}, DeepLabv3+ \cite{chen2018encoder}, DenseASPP~\cite{yang2018denseaspp}, UPerNet \cite{xiao2018unified}, DUpsampling \cite{tian2019decoders}, ESPNet \cite{mehta2018espnet} and GSCNN \cite{takikawa2019gated}) with eleven training data setups on our GMRP dataset, where the best result is highlighted in each subfigure.}
    \label{fig.onestream}
\end{figure*}

\begin{figure*}[!h]
    \centering
    \includegraphics[width=\textwidth]{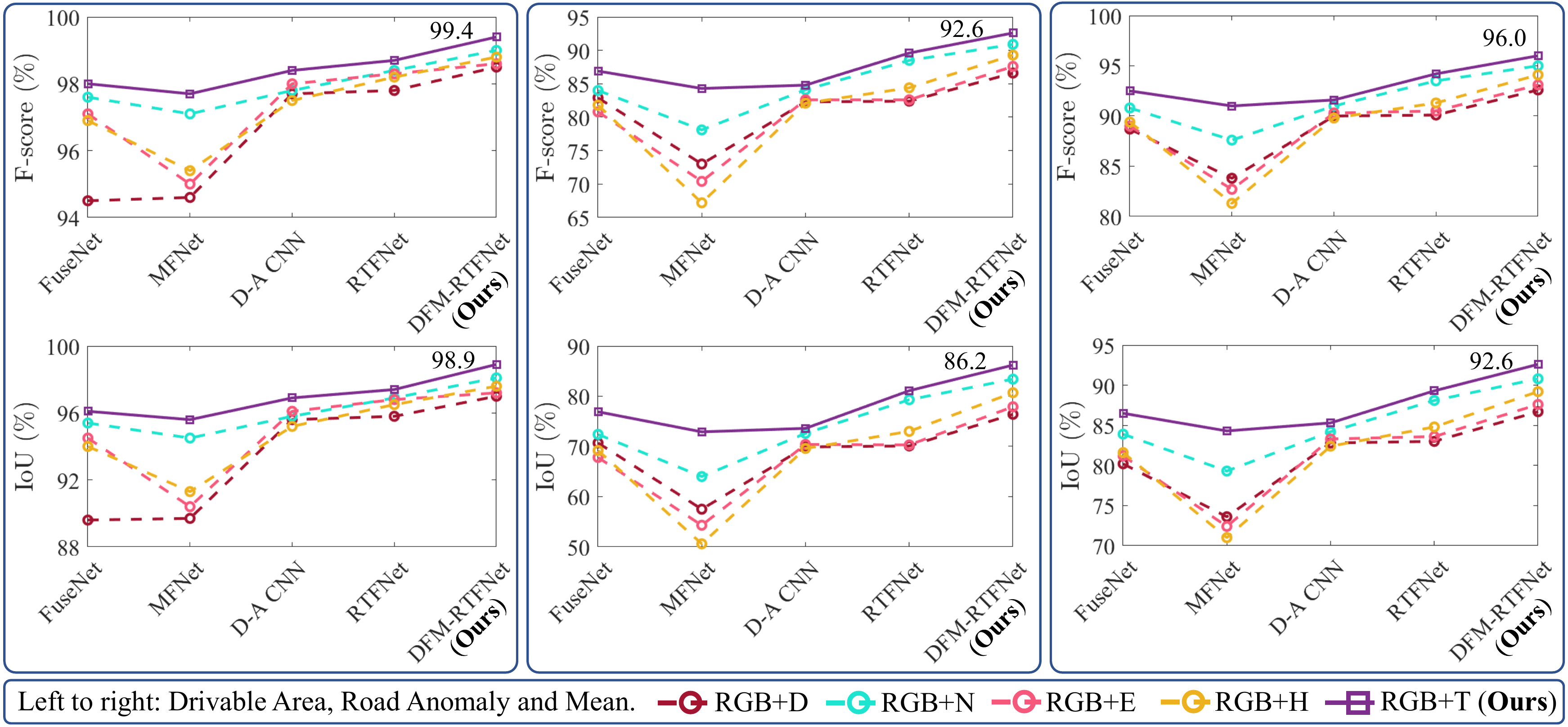}
    \caption{The performance comparison among four data-fusion networks (FuseNet \cite{hazirbas2016fusenet}, MFNet \cite{ha2017mfnet}, depth-aware CNN \cite{wang2018depth} and RTFNet \cite{sun2019rtfnet}) and our DFM-RTFNet with five training data setups on our GMRP dataset, where D-A CNN is short for depth-aware CNN, and the best result is highlighted in each subfigure.}
    \label{fig.twostream}
\end{figure*}

\section{Experimental Results and Discussion}
\label{sec.experiment_results_and_discussion}

\subsection{Datasets and Experimental Setups}
\label{sec.datasets_and_experimental_setups}
We utilize the following datasets in our experiments to evaluate the performance of our approach:
\begin{itemize}
    \item Our GMRP dataset \cite{wang2019self}: Our dataset is designed for ground mobile robots, and contains 3896 pairs of images with ground truth for drivable areas and road anomalies.
    \item The KITTI road dataset \cite{fritsch2013new}: This dataset is designed for self-driving cars, and contains 289 pairs of training data with ground truth for drivable areas and 290 pairs of testing data without ground truth.
    \item The KITTI semantic segmentation dataset \cite{alhaija2018augmented}: This dataset is also designed for self-driving cars, and contains 200 pairs of training data with ground truth for scene understanding and 200 pairs of testing data without ground truth.
\end{itemize}

The total 3896 pairs of images in our GMRP dataset are split into a training, a validation and a testing set that contain 2726, 585 and 585 pairs, respectively. The resolution of input images is downsampled to 320 $\times$ 480. We first conduct ablation studies on our GMRP dataset to (a) select the fusion strategy of our DFM-RTFNet, and (b) demonstrate the effectiveness and efficiency of our DFM, as presented in Section~\ref{sec.ablation_study}.

Then, Section~\ref{sec.drivable_area_and_road_anomaly_detection_benchmark} presents a drivable area and road anomaly detection benchmark for ground mobile robots, which provides a detailed performance comparison of the fourteen SOTA networks (ten single-modal ones and four data-fusion ones) mentioned in Section~\ref{sec.soa_networks_for_semantic_segmentation} and our DFM-RTFNet, with respect to six different types of training data, including our transformed disparity images. We train each single-modal network with eleven setups. Specifically, we first train each with input RGB, disparity, normal, elevation, HHA and transformed disparity images (denoted as \textbf{RGB}, \textbf{Disparity}, \textbf{Normal}, \textbf{Elevation}, \textbf{HHA} and \textbf{T-Disp}), respectively. Then we train each with input concatenation of RGB and the other five types of training data separately, denoted as \textbf{RGB+D}, \textbf{RGB+N}, \textbf{RGB+E}, \textbf{RGB+H} and \textbf{RGB+T}, followed by training each data-fusion network with same five setups. For the training setup, we use the stochastic gradient descent (SGD) optimizer. Moreover, we train each network until loss convergence and then select the best model according to the performance of the validation set. For the quantitative evaluations, we adopt the F-score (Fsc) and the Intersection over Union (IoU) for each class. We also plot the precision-recall curves and compute the average precision (AP) \cite{everingham2015pascal} for each class. Furthermore, we compute the mean values across all classes for the three metrics, denoted as mFsc, mIoU and mAP.

\begin{table}[t]
    \caption{The Experimental Results of Our Ablation Studies, where (A) is the Original RTFNet50 (the Baseline Setup); (B) is the Setup with Feature Concatenation; (C)--(E) are Three Setups with Different Fusion Strategies of Our DFM-RTFNet; (F) is the Setup with An SOTA Fusion Strategy; and (G) is the Original RTFNet101. Best Results are Bolded}
    \centering
    \begin{tabular}{lcccc}
        \toprule
        \multicolumn{1}{l}{Backbone} & \makecell{Fusion\\Strategy} & \makecell{mIoU\\($\%$)} & \makecell{Runtime\\(ms)} & \makecell{$\eta$\\($\%/$ms)} \\ \midrule
        (A) RTFNet50 & Addition & 89.3 & \textbf{24.7} & -- \\ \midrule
        (B) RTFNet50 & Concatenation & 88.6 & 25.3 & -1.17 \\
        (C) RTFNet50 & First DFM & 89.7 & 25.9 & 0.33 \\
        (D) RTFNet50 & Last DFM & 90.2 & 26.4 & 0.53 \\
        (E) RTFNet50 & All DFMs & \textbf{92.6} & 28.1 & \textbf{0.97} \\ \midrule
        (F) RTFNet50 & D-A Operators \cite{wang2018depth} & 90.8 & 27.6 & 0.52 \\ \midrule
        (G) RTFNet101 & Addition & 91.3 & 31.2 & 0.31 \\ \bottomrule
    \end{tabular}
    \label{tab.ablation_study}
\end{table}

To better understand how our transformed disparity image improves the overall detection performance, we compare it with another two visual features that can also make the drivable areas possess similar values. Additionally, we analyze the feature variation with and without our DFM to explore its internal mechanism for improving the detection performance. The experimental results are presented in Section~\ref{sec.further_discussion}.

Finally, we conduct experiments to demonstrate the effectiveness and efficiency of our approach for self-driving cars. Since our drivable area detection task perfectly matches the KITTI road benchmark \cite{fritsch2013new}, we train our DFM-RTFNet using the KITTI road training data and submit it to the benchmark. However, since we focus on the detection of drivable areas and road anomalies, we do not submit the results of our approach to the KITTI semantic segmentation benchmark \cite{alhaija2018augmented}. Instead, we merge its classes into four new classes, unlabeled, drivable area, vehicles and pedestrians, because vehicles and pedestrians are two important anomalies for self-driving cars. We split the KITTI semantic segmentation training data into a training, a validation and a test set that contains 100, 50 and 50 pairs of data, respectively. Then, we compare the performances between our DFM-RTFNet and four SOTA data-fusion networks with respect to six different types of training data. The experimental results are presented in Section~\ref{sec.evaluations_on_the_kitti_benchmark}.

\subsection{Ablation Study}
\label{sec.ablation_study}
In this subsection, we adopt RTFNet50 \cite{sun2019rtfnet} with input transformed disparity images as the baseline to conduct ablation studies, and (A) of Tab.~\ref{tab.ablation_study} shows the performance of the baseline. To compare the differences between the setups more intuitively, we introduce two new metrics: (a) the runtime of a given setup on an NVIDIA GeForce GTX 1080 Ti graphics card; and (b) the ratio of the mIoU increment and runtime increment between a given setup and the baseline $\eta$. Let $\eta_i$ denote the $\eta$ of a given setup $i$, and the expression of $\eta_i$ is
\begin{equation}
    \eta_i = \frac{\text{mIoU}_{i}-\text{mIoU}_\text{baseline}}{\text{Runtime}_{i}-\text{Runtime}_\text{baseline}} (\%/ms).
\end{equation}
$\eta$ is used to quantify the trade-off between the improved performance and increased runtime of a given setup. An effective and efficient setup achieves a high $\eta$ value.

\begin{table}[t]
    \caption{The Performance Comparison ($\%$) on our GMRP Dataset, where $\text{AP}_{D}$ and $\text{AP}_{R}$ denote the AP for drivable areas and road anomalies, respectively. Best Results are Bolded}
    \centering
    \begin{tabular}{L{2.1cm}C{1.9cm}C{0.8cm}C{0.8cm}C{0.8cm}}
        \toprule
        \multicolumn{1}{l}{Approach} & Setup & $\text{AP}_{D}$ & $\text{AP}_{R}$ & mAP \\ \midrule
        DeepLabv3+ \cite{chen2018encoder} & T-Disp (\textbf{Ours}) & 99.71 & 92.45 & 96.08 \\
        ESPNet \cite{mehta2018espnet} & T-Disp (\textbf{Ours}) & 99.68 & 91.79 & 95.74 \\
        GSCNN \cite{takikawa2019gated} & T-Disp (\textbf{Ours}) & 99.36 & 93.61 & 96.49 \\ \midrule
        FuseNet \cite{hazirbas2016fusenet} & RGB+T (\textbf{Ours}) & 99.25 & 93.39 & 96.32 \\
        RTFNet \cite{sun2019rtfnet} & RGB+T (\textbf{Ours}) & 99.70 & 96.27 & 97.99 \\ \midrule
        \multirow{5}{*}{\makecell{DFM-RTFNet\\(\textbf{Ours})}} & RGB+D & 99.72 & 92.17 & 95.95 \\
        & RGB+N & 99.67 & 97.12 & 98.40 \\
        & RGB+E & 99.69 & 94.83 & 97.26 \\
        & RGB+H & 99.61 & 96.13 & 97.87 \\
        & RGB+T (\textbf{Ours}) & \textbf{99.85} & \textbf{97.61} & \textbf{98.73} \\ \bottomrule
    \end{tabular}
    \label{tab.AP}
\end{table}

We first explore different fusion strategies of our DFM-RTFNet and the corresponding performance is presented in (C)--(E) of Tab.~\ref{tab.ablation_study}. (C) and (D) mean that we only replace the first and last element-wise addition layer, respectively, with our DFM, and (E) represents the setup shown in Fig.~\ref{fig.network}. We can observe that (C)--(E) all outperform the commonly used element-wise addition and feature concatenation strategies shown in (A) and (B) of Tab.~\ref{tab.ablation_study}, which demonstrates that our DFM is an effective module for data fusion. Furthermore, (E) presents the best performance, and therefore, we adopt the setup illustrated in Fig.~\ref{fig.network} in the rest of our experiments.

In addition, (F) of Tab.~\ref{tab.ablation_study} presents the performance of the setup with an SOTA fusion strategy, depth-aware (D-A) operators \cite{wang2018depth}. We can see that our DFM outperforms it with a higher $\eta$ value. Moreover, one exciting fact is that our DFM-RTFNet with the backbone of RTFNet50 ((E) of Tab.~\ref{tab.ablation_study}) even presents a better performance than the original RTFNet101 \cite{sun2019rtfnet} ((G) of Tab.~\ref{tab.ablation_study}) and the runtime is much less, which demonstrates the effectiveness and efficiency of our DFM.

\subsection{Drivable Area and Road Anomaly Detection Benchmark}
\label{sec.drivable_area_and_road_anomaly_detection_benchmark}
\begin{figure}[t]
    \centering
    \includegraphics[width=\linewidth]{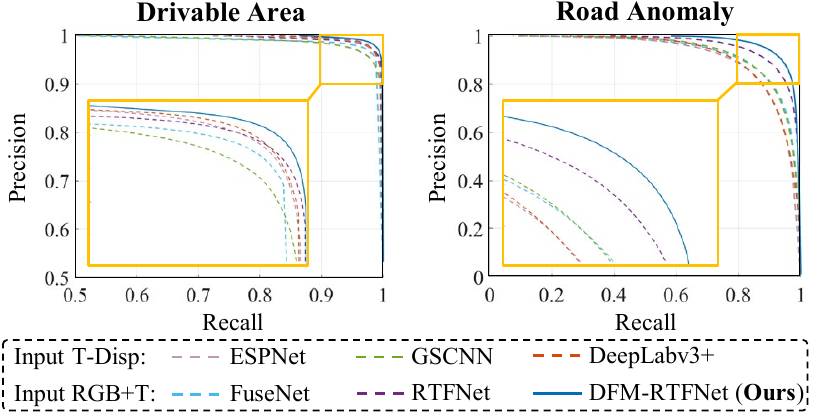}
    \caption{The precision-recall curves of different networks on our GMRP dataset. ESPNet \cite{mehta2018espnet}, GSCNN \cite{takikawa2019gated} and DeepLabv3+ \cite{chen2018encoder} are single-modal networks, while FuseNet \cite{hazirbas2016fusenet}, RTFNet \cite{sun2019rtfnet} and our DFM-RTFNet are data-fusion networks. The orange boxes show the enlarged area for comparison.}
    \label{fig.PR_network}
\end{figure}

\begin{figure}[t]
    \centering
    \includegraphics[width=\linewidth]{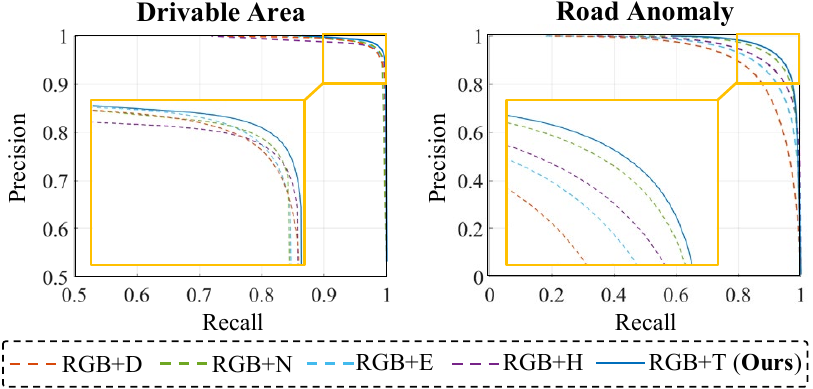}
    \caption{The precision-recall curves of our DFM-RTFNet with five training data setups on our GMRP dataset. The orange boxes show the enlarged area for comparison.}
    \label{fig.PR_input}
\end{figure}

In this subsection, we present a drivable area and road anomaly detection benchmark for ground mobile robots. Fig.~\ref{fig.benchmark} presents sample qualitative results, from which we can find that our transformed disparity images greatly help reduce the noises in the semantic predictions. Moreover, our DFM-RTFNet presents more accurate and robust results than all the other SOTA data-fusion networks with the same input. The corresponding quantitative performances are presented in Fig.~\ref{fig.onestream} and Fig.~\ref{fig.twostream}, and the detailed numbers are provided in the benchmark page\footnote{\url{https://sites.google.com/view/gmrb}} for reference. Specifically, our DFM-RTFNet with our transformed disparity images as input increases the mean F-score and mean IoU by around 1.0--21.7$\%$ and 1.8--31.9$\%$, respectively. We also select several networks that perform well for further performance comparison. The AP comparison is presented in Tab.~\ref{tab.AP}, and the precision-recall curves are shown in Fig.~\ref{fig.PR_network} and Fig.~\ref{fig.PR_input}. We can clearly observe that our DFM-RTFNet with our transformed disparity images as input presents the best performance. This proves that our transformed disparity images and DFM can effectively improve the detection performance. In addition, we conduct more experiments to explore how our proposed approach enhances the detection performance, and the experimental results are presented in Section~\ref{sec.further_discussion}.

\subsection{Further Discussion}
\label{sec.further_discussion}

\begin{figure}[t]
    \centering
    \includegraphics[width=\linewidth]{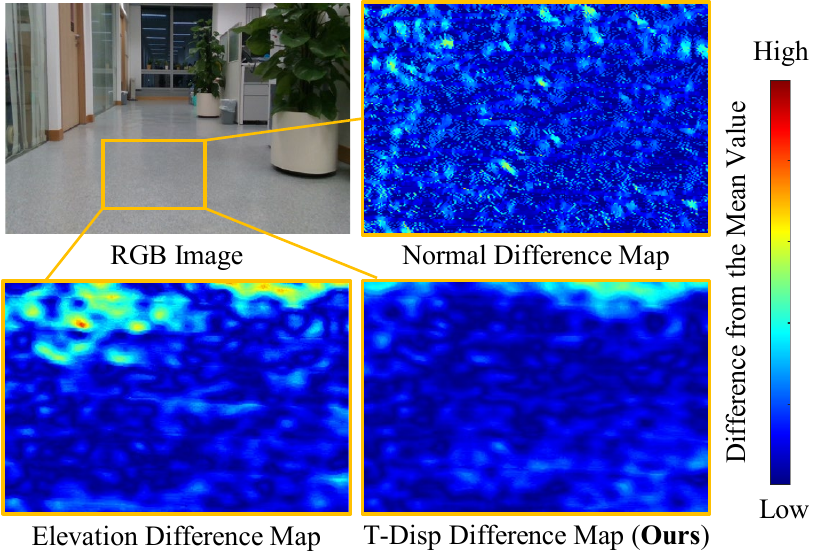}
    \caption{An example of the drivable area consistency comparison between normal images, elevation maps and our transformed disparity images. The orange box in the RGB image shows the enlarged area for comparison, and the other three images present the difference maps from the corresponding mean values in the enlarged area.}
    \label{fig.prove_TD}
\end{figure}

As aforementioned, our transformed disparity images can make drivable areas possess similar values, and thus improve the detection performance. To verify this, we compare our transformed disparity images with the other two visual features that have similar properties, \ie, normal images and elevation maps. Since these three features have different scales, we introduce a dimensionless metric, the coefficient of variation $c_v$, defined as follows:
\begin{equation}
    c_v = \frac{\sigma}{\mu},
\end{equation}
where $\sigma$ and $\mu$ denote the standard deviation and mean, respectively. A uniform data distribution achieves a low $c_v$ value. Then, we compute the $c_v$ of the three visual features on the drivable areas in our GMRP dataset. Note that for three-channel normal images, we first compute the average values across all channels and use the obtained one-channel average maps for comparison. The $c_v$ values of normal images~\cite{badino2011fast}, elevation maps \cite{gupta2014learning} and our transformed disparity images are 0.008, 0.009 and 0.005, respectively. We can see that our transformed disparity images achieve a much lower $c_v$ value. The sample qualitative results shown in Fig.~\ref{fig.prove_TD} also verify our conclusion that our transformed disparity images can make the drivable areas more uniform and thus benefit all networks for better detection performances.

\begin{table}[t]
    \caption{The Experimental Results ($\%$) of Two SOTA Data-Fusion Networks with and without Our DFM on Our GMRP Dataset. Best Results for Each Network are Bolded}
    \centering
    \begin{tabular}{L{1.9cm}C{0.65cm}C{0.65cm}C{0.65cm}C{0.65cm}C{0.65cm}C{0.65cm}}
        \toprule
        \multicolumn{1}{l}{\multirow{2}{*}{Approach}} & \multicolumn{2}{c}{Drivable Area} & \multicolumn{2}{c}{Road Anomaly} & \multicolumn{2}{c}{Mean} \\ \cmidrule(l){2-3} \cmidrule(l){4-5} \cmidrule(l){6-7}
        \multicolumn{1}{c}{} & Fsc & IoU & Fsc & IoU & Fsc & IoU \\ \midrule
        FuseNet \cite{hazirbas2016fusenet} & 98.0 & 96.1 & 86.9 & 76.9 & 92.5 & 86.5 \\
        DFM-FuseNet & \textbf{98.6} & \textbf{97.3} & \textbf{90.6} & \textbf{82.8} & \textbf{94.6} & \textbf{90.1} \\ \midrule
        RTFNet \cite{sun2019rtfnet} & 98.7 & 97.4 & 89.6 & 81.1 & 94.2 & 89.3 \\
        DFM-RTFNet & \textbf{99.4} & \textbf{98.9} & \textbf{92.6} & \textbf{86.2} & \textbf{96.0} & \textbf{92.6} \\ \bottomrule
    \end{tabular}
    \label{tab.prove_DFM}
\end{table}

\begin{figure*}[t]
    \centering
    \includegraphics[width=\textwidth]{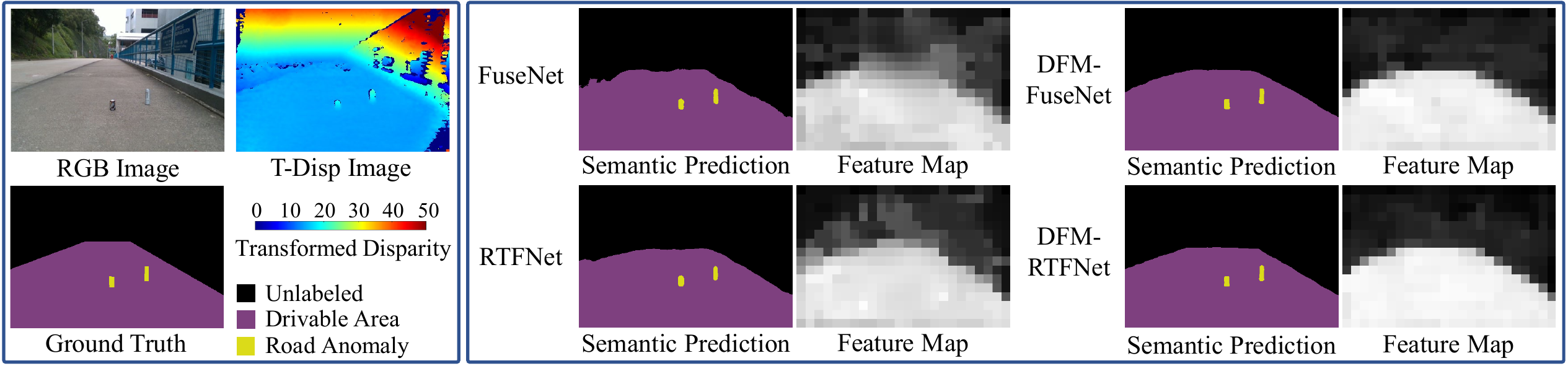}
    \caption{An example of two SOTA data-fusion networks (FuseNet \cite{hazirbas2016fusenet} and RTFNet \cite{sun2019rtfnet}) with and without our DFM embedded, respectively. Feature maps refer to the mean activation maps of the features after the last layers of the encoders.}
    \label{fig.prove_DFM}
\end{figure*}

To explore the internal mechanism of our DFM for improving the detection performance, we implement it in FuseNet \cite{hazirbas2016fusenet}, besides RTFNet \cite{sun2019rtfnet}, with the RGB+T setup on our GMRP dataset. The quantitative comparisons are given in Tab.~\ref{tab.prove_DFM}, where we can observe that networks with our DFM embedded generally perform better than themselves without our DFM embedded. Moreover, we visualize the mean activation maps of the features after the last layers of the encoders with and without our DFM for each network, as presented in Fig.~\ref{fig.prove_DFM}. We can observe that the mean activation maps fused by our DFM can concentrate more on the drivable areas and road anomalies. We conjecture the reason is that the content-dependent and spatially variant properties make our DFM act as a weight modulation operator, which can effectively generate fused features with high activation values in the critical areas, and thus improve the detection performance.

\subsection{Evaluations on the KITTI Benchmark}
\label{sec.evaluations_on_the_kitti_benchmark}
\subsubsection{KITTI Road Benchmark}
\begin{figure*}[t]
    \centering
    \includegraphics[width=\textwidth]{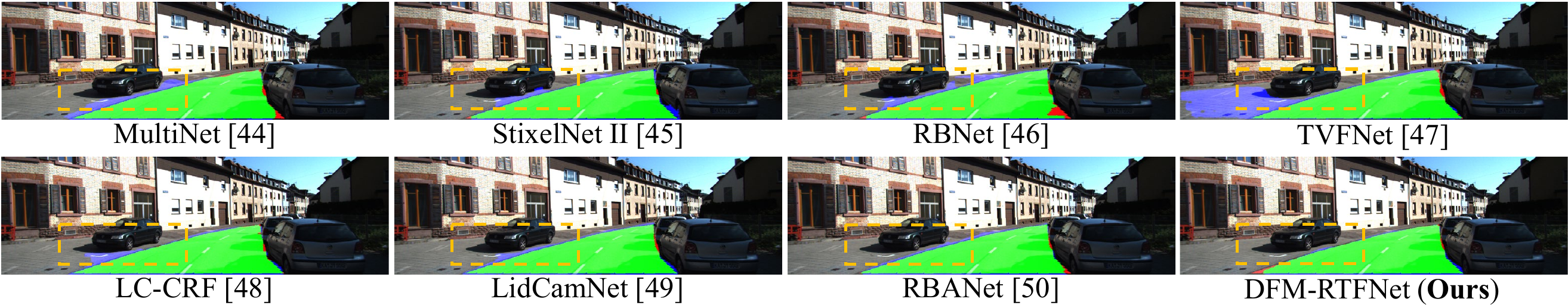}
    \caption{An example of the testing images on the KITTI road benchmark. The true positive, false negative and false positive pixels are shown in green, red and blue, respectively. Significantly improved regions are marked with orange dashed boxes.}
    \label{fig.kitti_road}
\end{figure*}

As previously mentioned, we submit the road estimation results of our DFM-RTFNet to the KITTI road benchmark \cite{fritsch2013new}. Experimental results demonstrate that our DFM-RTFNet achieves competitive performance on the benchmark. Fig.~\ref{fig.kitti_road} illustrates an example of the testing images, and Tab.~\ref{tab.kitti_road} presents the quantitative results. We can observe that our DFM-RTFNet outperforms existing approaches, which demonstrates its effectiveness and efficiency for self-driving cars.

\subsubsection{KITTI Semantic Segmentation Dataset}
\begin{figure*}[t]
    \centering
    \includegraphics[width=\textwidth]{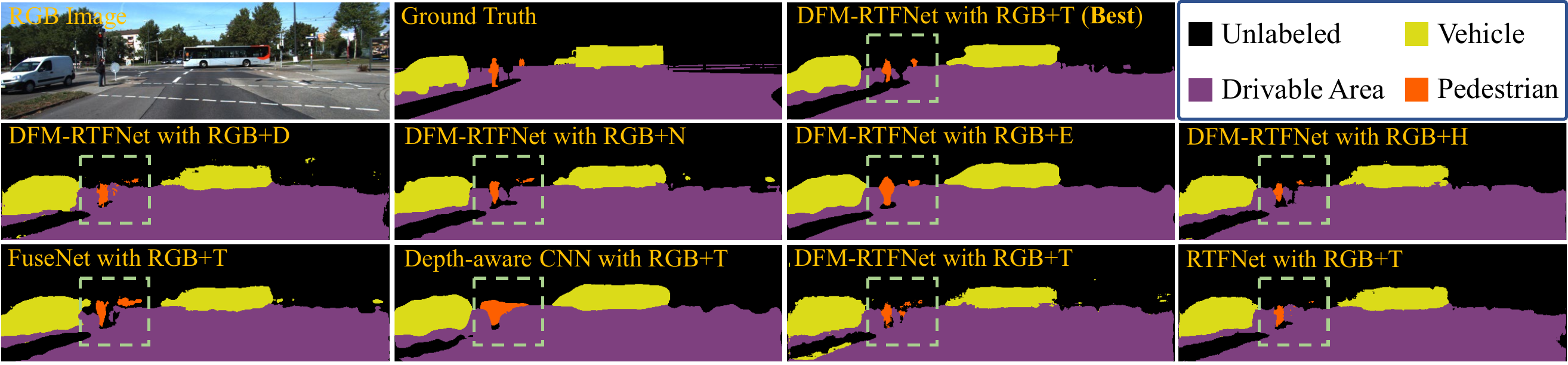}
    \caption{An example of the experimental results on the KITTI semantic segmentation dataset. FuseNet \cite{hazirbas2016fusenet}, MFNet \cite{ha2017mfnet}, Depth-aware CNN \cite{wang2018depth}, RTFNet~\cite{sun2019rtfnet} and our DFM-RTFNet are all data-fusion networks. Significantly improved regions are marked with green dashed boxes.}
    \label{fig.kitti_segmentation}
\end{figure*}

Fig.~\ref{fig.kitti_segmentation} and Tab.~\ref{tab.kitti_segmentation} present the qualitative and quantitative results on the KITTI semantic segmentation dataset \cite{alhaija2018augmented}, respectively. We can see that our transformed disparity images greatly improve the detection performance, and our DFM-RTFNet presents a better performance than all other SOTA data-fusion networks, which verifies that our DFM-RTFNet with the RGB+T setup can be deployed effectively for self-driving cars in practice.

\begin{table}[t]
    \caption{KITTI Road Benchmark$^3$ Results, Where the Best Results are Bolded. Note that We Only Compare Our Approach with Other Published Approaches}
    \centering
    \begin{tabular}{L{2.8cm}C{1.2cm}C{1.2cm}C{1.4cm}}
        \toprule
        \multicolumn{1}{l}{Approach} & MaxF ($\%$) & AP ($\%$) & Runtime (s) \\ \midrule
        MultiNet \cite{teichmann2018multinet} & 94.88 & 93.71 & 0.17 \\
        StixelNet II \cite{garnett2017real} & 94.88 & 87.75 & 1.20 \\
        RBNet \cite{chen2017rbnet} & 94.97 & 91.49 & 0.18 \\
        TVFNet \cite{gu2019two} & 95.34 & 90.26 & \textbf{0.04} \\
        LC-CRF \cite{gu2019road} & 95.68 & 88.34 & 0.18 \\
        LidCamNet \cite{caltagirone2019lidar} & 96.03 & 93.93 & 0.15 \\
        RBANet \cite{sun2019reverse} & 96.30 & 89.72 & 0.16 \\ \midrule
        DFM-RTFNet (\textbf{Ours}) & \textbf{96.78} & \textbf{94.05} & 0.08 \\ \bottomrule
    \end{tabular}
    \label{tab.kitti_road}
\end{table}

\footnotetext[3]{\url{http://www.cvlibs.net/datasets/kitti/eval_road.php}}

\section{Conclusion}
\label{sec.conclusion_and_future_work}
In this paper, we conducted a comprehensive study on drivable area and road anomaly detection for mobile robots, including building a benchmark and proposing an effective and efficient data-fusion strategy named DFM. Experimental results verified that our transformed disparity images could enable drivable areas exhibit similar values, which could benefit networks for drivable area and road anomaly detection. Moreover, our proposed DFM can effectively generate fused features with high activation values in critical areas using content-dependent and spatially variant kernels, and thus improve the overall detection performance. Compared with the SOTA networks, our DFM-RTFNet can produce more accurate and robust results for drivable area and road anomaly detection. Hence, it is suitable to be implemented in mobile robots in practice. Furthermore, our DFM-RTFNet achieves competitive performance on the KITTI road benchmark. We believe that our benchmark and the data fusion idea in our proposed network will inspire future research in this area. In the future, we will develop effective and efficient data-fusion strategies to further improve the detection performance.

\begin{table}[t]
    \caption{The Experimental Results ($\%$) of Four SOTA Data-Fusion Networks and Our DFM-RTFNet with Respect to Different Training Data Setups on the KITTI Semantic Segmentation Dataset. Best Results are Bolded}
    \centering
    \begin{tabular}{L{2.7cm}C{2cm}C{1cm}C{1cm}}
        \toprule
        \multicolumn{1}{l}{Approach} & Setup & mFsc & mIoU \\ \midrule
        FuseNet \cite{hazirbas2016fusenet}  & RGB+T (\textbf{Ours}) & 88.4 & 79.6 \\
        MFNet \cite{ha2017mfnet} & RGB+T (\textbf{Ours}) & 85.5 & 75.2 \\
        Depth-aware CNN \cite{wang2018depth} & RGB+T (\textbf{Ours}) & 87.0 & 77.5 \\
        RTFNet \cite{sun2019rtfnet} & RGB+T (\textbf{Ours}) & 90.3 & 82.7 \\ \midrule
        \multirow{5}{*}{DFM-RTFNet (\textbf{Ours})} & RGB+D & 88.2 & 79.5 \\
        & RGB+N & 90.6 & 83.2 \\
        & RGB+E & 89.3 & 81.2 \\
        & RGB+H & 89.8 & 82.1 \\
        & RGB+T (\textbf{Ours}) & \textbf{92.5} & \textbf{86.5} \\ \bottomrule
    \end{tabular}
    \label{tab.kitti_segmentation}
\end{table}

\bibliographystyle{IEEEtran}
\bibliography{ref}

\end{document}